%% file: paper.tex
\documentclass[sn-mathphys,Numbered,pdflatex,iicol]{sn-jnl}

\usepackage{amssymb}
\usepackage[T1]{fontenc}
\usepackage{lmodern}
\usepackage{microtype}
\usepackage{xcolor}
\usepackage{xspace}
\usepackage{booktabs}
\usepackage{caption}
\usepackage{subcaption}

\usepackage{graphicx}%
\usepackage{multirow}%
\usepackage{amsmath,amssymb,amsfonts}%
\usepackage{amsthm}%
\usepackage{mathrsfs}%
\usepackage[title]{appendix}%
\usepackage{xcolor}%
\usepackage{textcomp}%
\usepackage{manyfoot}%
\usepackage{booktabs}%
\usepackage{algorithm}%
\usepackage{algorithmicx}%
\usepackage{algpseudocode}%
\usepackage{listings}%

\usepackage{algorithm}
\usepackage{algpseudocode}

\usepackage{dblfloatfix}
\usepackage{placeins}

\usepackage{orcidlink}

\newcommand{\IoU}{\ensuremath{I \mkern-2mu o \mkern1mu U\mkern-1mu}\xspace}

\begin{document}
	
	\title{Uncertainty estimates for semantic segmentation: providing enhanced reliability for automated motor claims handling}

\author*[1]{\fnm{Jan} \sur{Küchler} \orcidlink{0000-0001-9087-6230}}\email{j.kuechler@controlexpert.com}
\author[1,a]{\fnm{Daniel} \sur{Kröll}}\email{d.kroell@controlexpert.com}
\author[1,a]{\fnm{Sebastian}
\sur{Schoenen}}\email{s.schoenen@controlexpert.com}
\author[1,a]{\fnm{Andreas}
\sur{Witte}}\email{a.witte@controlexpert.com}

\affil[1]{\orgname{ControlExpert GmbH}, \orgaddress{\street{Marie-Curie-Straße 3}, \city{Langenfeld}, \postcode{40764}, \country{Germany}}}
\affil[a]{These authors contributed equally to this work}

	\date{\today}

	\abstract{
		Deep neural network models for image segmentation can be a powerful tool for the automation of motor claims handling processes in the insurance industry. A crucial aspect is the reliability of the model outputs when facing adverse conditions, 
		such as low quality photos taken by claimants to document damages.
		 We explore the use of a meta-classification model to empirically assess the precision of segments predicted by a  model trained for the semantic segmentation of car  body parts. Different sets of features correlated with the quality of a segment are compared, and an AUROC score of 0.915 is achieved for distinguishing between high- and low-quality segments. By removing low-quality segments, the average $m\IoU$ of the segmentation output is improved by 16 percentage points and the number of wrongly predicted segments is reduced by 77\%.
	}


\maketitle

\section{Introduction}
	
In the rapidly evolving world of automotive insurance, technological advancements are reshaping the landscape. Efficient and accurate claims handling remain key success factors for the insurance industry. At the heart of this process is damage assessment, traditionally reliant on manual methods. This procedure often required experts to either make on-site visits to inspect damaged cars or, increasingly common today, review photographs provided by claimants. While this approach is thorough, it is also time-consuming and vulnerable to human biases and errors.

\begin{figure*}[tb]
	\includegraphics[width=\columnwidth]{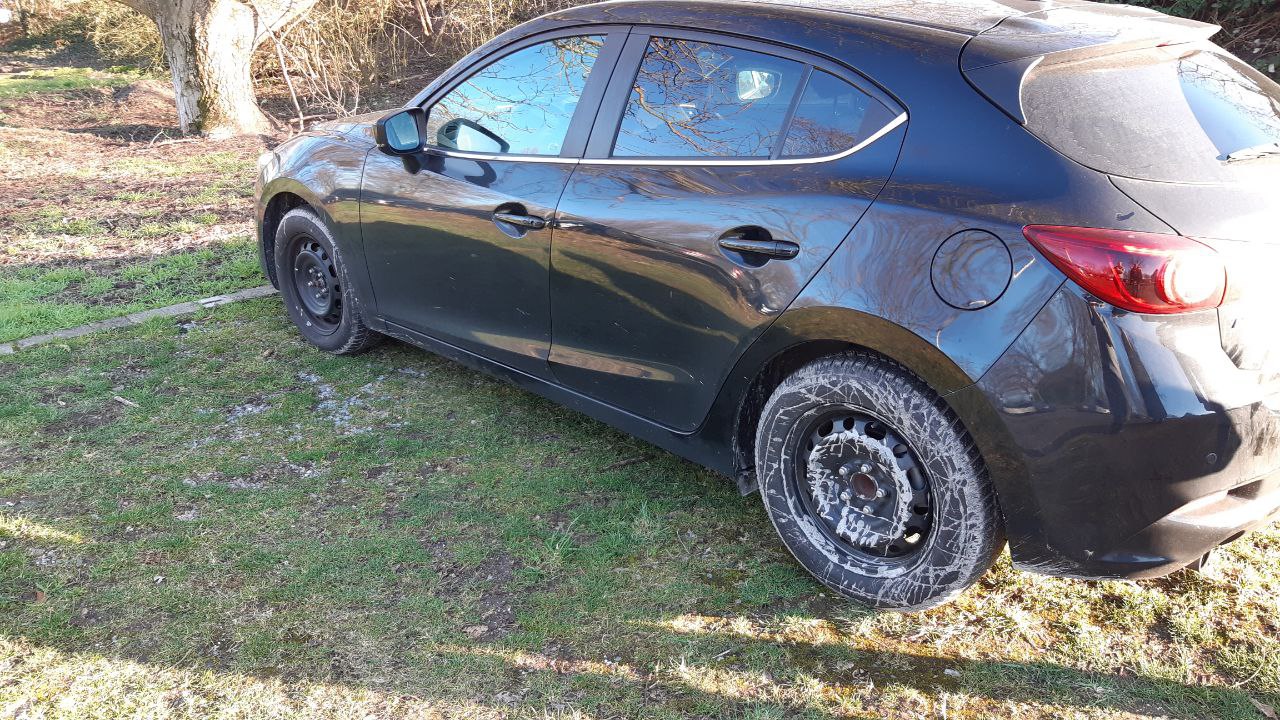}%
	\hfill%
	\includegraphics[width=\columnwidth]{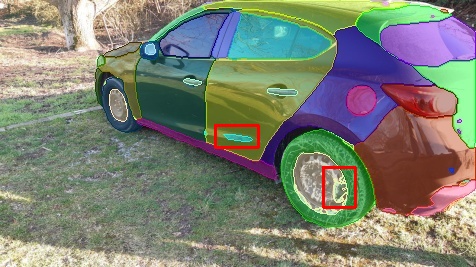}
	
	\caption{Photograph of a car (left), taken to highlight issues that can negatively affect the performance of DNN segmentation models: reflections, dirt and bad exposure. Result of our semantic segmentation model (right), trained to segment car body parts. Predicted segments are shown as colored overlays. A few mistakes in the prediction are highlighted by red boxes: a reflection on the door is segmented as a molding part, and a part of the rear left rim is identified as an air intake.}
	\label{fig:example-car}
\end{figure*}

The advent of new computer vision techniques, particularly semantic segmentation~\cite{semantic-segmentation}, opens up possibilities to automate and streamline the damage assessment process. By segmenting images into categorized car parts and damages, this holds the potential to identify, classify and localize car damages. Embracing these techniques could empower the insurance industry to cut operational costs, expedite claim processing, and crucially, boost accuracy.

However, any technology-driven solution requires rigorous validation of its reliability. While deep neural networks (DNNs) have demonstrated exceptional performance in semantic segmentation tasks \cite{segmentation-hrnet, segmentation-review}, the variability in images of damaged cars -- influenced by factors like lighting conditions, vehicle models, capture angles, and other variables -- can introduce uncertainties. Addressing this challenge is of paramount importance. 

\figurename~\ref{fig:example-car} shows an example for an image with some of the aforementioned issues, together with the semantic segmentation mask of car body parts. Among other mistakes, a small area at the rim of the rear left wheel is identified as an air intake, likely due to dirt obscuring the usual features expected for a wheel.

To ensure a reliable and trustworthy damage assessment leveraging these technologies, the incorporation of uncertainty estimates into semantic segmentation is indispensable. By doing so, the industry can not only revolutionize the damage assessment but also make it transparent, consistent, and trustworthy, truly elevating the standards of automotive insurance claims handling.

\begin{figure*}[tb]
	\includegraphics[width=\linewidth]{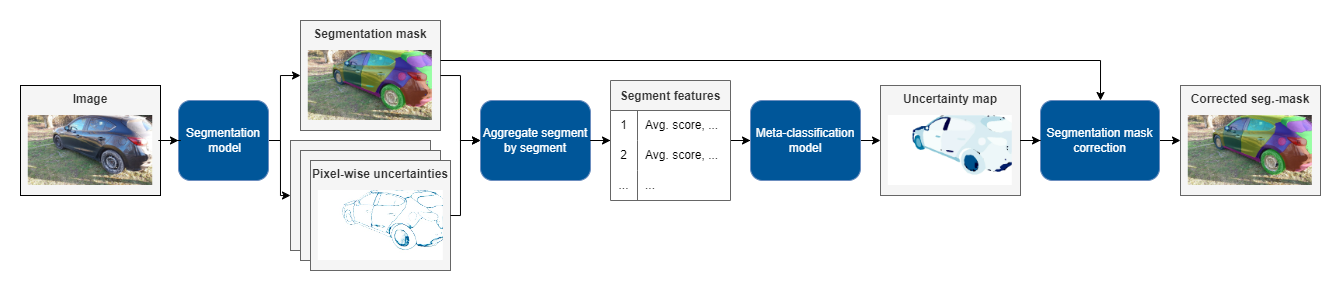}
	\caption{Schematic diagram of the explored method. An input image is processed by a semantic segmentation model and the resulting segmentation mask and softmax probabilities are aggregated to segment wise features. These are processed by a meta-classification model in order to produce a segment-wise uncertainty map. Finally, the segment uncertainties are used to correct the segmentation mask.}
	\label{fig:diagram}
\end{figure*}

Various approaches have been proposed to provide a measure of uncertainty in the model results for semantic segmentation. Modern architectures steadily improve the robustness of segmentation models, but they do not improve in terms of uncertainty estimation and calibration~\cite{reliabilitytrack}. While the output scores of a DNN are correlated with the accuracy of the result, models are often overconfident and output high probabilities even for wrong results \cite{score-confidence,relu-overconfidence,confidence-calibration}. 
In general, uncertainty quantification for deep learning is a widely studied topic~\cite{DeepLearningUQReview}, with techniques comprising primarily Bayesian approaches and ensemble methods, but also empirical methods to estimate uncertainties.
Monte Carlo dropout is used in a Bayesian framework to estimate model uncertainties~\cite{BayesianUNet,bayesian-segnet}, and can be combination with test-time image augmentation to also encompass data uncertainties~\cite{SegmentationDropoutAugmentation}. A technique called `Bayes by Backprop' is an alternative approach principled in the minimization of the variational free energy and used to quantify the uncertainty in the learned weights~\cite{BBB}. Ensemble methods assess the uncertainty by comparing the results of multiple, slightly different models trained for the same task \citep{ensemble}, and have been found to give a well calibrated result probability~\cite{SegmentationUncertaintyReview}. Using distillation techniques, even single models can be trained to predict the pixel-wise uncertainty in a segmentation result \citep{DUDESdistillation,distillation}, thus reducing the computational demands at inference time.

In this work, we explore the use of a meta-classification~\cite{metaclassification} model to empirically estimate the uncertainty of individual segments~\cite{metaseg}.
Although this approach is not based on a theoretical foundation, it has the advantage of neither requiring modifications to the segmentation model, nor to its training, and has a relatively low computational overhead during inference. 

As detailed in the following, uncertainty measures are first defined pixel by pixel, based on the softmax probability output of the segmentation network together with the loss gradient of the last convolutional layer. They are aggregated over predicted segments, and used, together with the predicted class of a segment and its size, to build a classification model that distinguishes between well and wrongly predicted segments. The score of this classifier is used as a measure of the uncertainty in the prediction. A low uncertainty result can be automatically processed with high confidence, while a high uncertainty score can indicate the need of human oversight. In special cases, the uncertainty score can be used to improve the segmentation mask. By removing segments with a high uncertainty from the segmentation mask, the precision of the segmentation output can be improved for the cost of reducing the recall. \figurename~\ref{fig:diagram} shows a schematic diagram of the method.

\section{Pixel- and segment-wise uncertainty measures}

The output of a semantic segmentation network with a final softmax layer are the pixel-wise probabilities $p_i^k$ for every semantic class $k=1, ..., N$, with the index $i$ running over all pixel coordinates. The predicted class for every pixel is the one with the highest probability, $\hat c_i = \arg\max_k p_i^k$.

The probability of the predicted class for a pixel, $\hat{p}_i = \max_k p_i^k$ quantifies the confidence in the result \citep{softmax-baseline}, thus $1-\hat{p}_i$ is used as one measure of the pixel-wise uncertainty.

Following\footnote{\citeauthor{metaseg} make the source code of their method available at \url{https://github.com/mrottmann/MetaSeg}} \cite{metaseg}, two further quantities are defined, measuring the dispersion of the pixel-wise probabilities:
\begin{itemize}
	\item the entropy
		\[
			E_i = \frac{1}{\log K} \sum_{k=1}^N p_i^k \log(p_i^k)\ ,
		\]
		which is maximized when the model result sees all classes as equally likely,
	\item as well as the difference between the two largest softmax values, 
		\[
			D_i = \hat{p}_i - \max_{k \neq \hat c_i} p_i^k\ ,
		\]
		which targets cases where the network predicts a similar probability for the two most likely classes.
\end{itemize} 

In \cite{gradient-uncertainty}, a gradient-based approach for uncertainty quantification in semantic segmentation is introduced. The gradient of a categorical cross entropy loss with respect to the last convolutional layer of the segmentation network can be computed efficiently. When taking the predicted class $\hat{c_i}$ as the one-hot label per pixel, these gradients quantify how similar the result is to the examples in the training data set. Intuitively, larger gradients mean that the weights of the convolutional layer need to be changed more strongly to accommodate the input, therefore indicating an uncertain result. The norm of the pixel-wise gradients is taken as an additional measure of the uncertainty, which can be efficiently computed~\cite{gradient-uncertainty} as $G_i = \left\|p_i^k (1-\delta_{k\hat{c_i}}) \psi_i\right\|_2$, with $\psi_i$ denoting the features before the last convolution layer.

\begin{figure}[tb]
	\includegraphics[width=.49\linewidth]{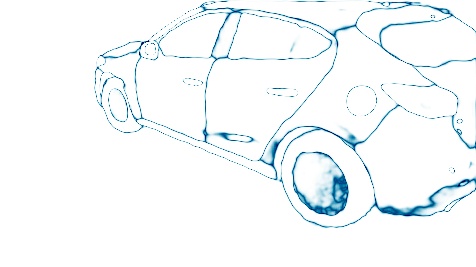}%
	\hfill%
	\includegraphics[width=.49\linewidth]{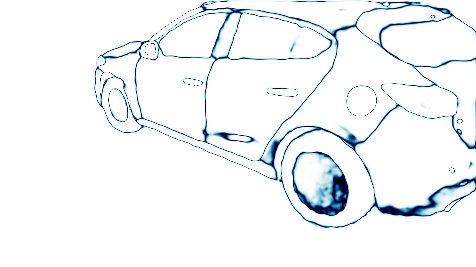}
	
	\vspace{.02\linewidth}
	\includegraphics[width=.49\linewidth]{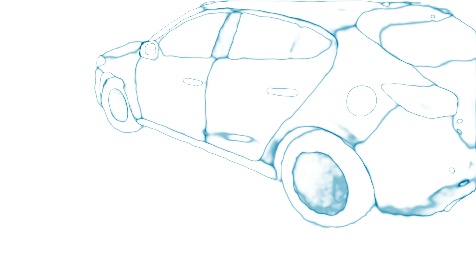}%
	\hfill%
	\includegraphics[width=.49\linewidth]{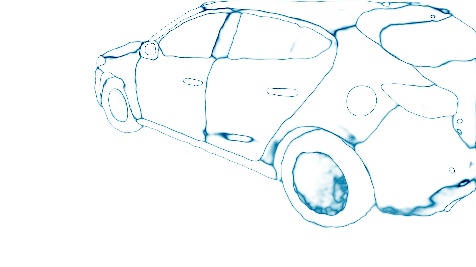}
	\caption{Qualitative heat-maps of $1-\hat{p}_i$ (top left), $1 - D_i$ (top right), the entropy $E_i$ (bottom left) and the gradient uncertainty $G_i$ (bottom right), for the example image shown in \figurename~\ref{fig:example-car}. Darker shades indicate higher pixel-wise uncertainties.}
	\label{fig:heatmaps}
\end{figure}

\figurename~\ref{fig:heatmaps} shows qualitative heat-maps of the pixel-wise uncertainty measures for the example image of \figurename~\ref{fig:example-car}. Due to the labeling accuracy, the boundaries between segments of different classes are uncertain and highlighted in the heat-maps. The wrongly predicted segments at the door and at the rim of the rear left wheel are also indicated by high pixel-wise uncertainties.
On the other hand, the uncertainties vary strongly in these segments. 
The pixel-wise uncertainties are aggregated to segment-wise measures, in order to build features for the classification of high- and low-quality segments. The aggregation of uncertainty estimates from pixel to segment level has been shown to improve the performance for the detection of anomalies by accounting for the correlation between neighboring pixels~\cite{MaskLevelAdvantages}.

The predicted semantic segmentation mask for an image is split into a set $\hat{\mathcal{K}}$ of segments, i.e. connected areas of the same class.
Segment by segment, the pixel-wise uncertainty measures are averaged over all pixels of the segment, e.g. the mean entropy $E(\hat{k})$ of a segment $\hat{k}\in\hat{\mathcal{K}}$ is $E(\hat{k}) = 1/|\hat{k}| \sum_{i\in \hat{k}} E_i$ and analogously for the other uncertainty measures. 
The values are also averaged separately over the boundary and the inner part of the segment, as defined by~\cite{metaseg}, because the boundaries typically exhibit higher uncertainties. Additionally, the standard deviation of the pixel-wise uncertainty distributions on the boundary, inner and full segment is used as an input to the meta-classification model.

\begin{figure*}[tbp]
	\centering

	\begin{subfigure}{.6\textwidth}
		\includegraphics[width=\linewidth]{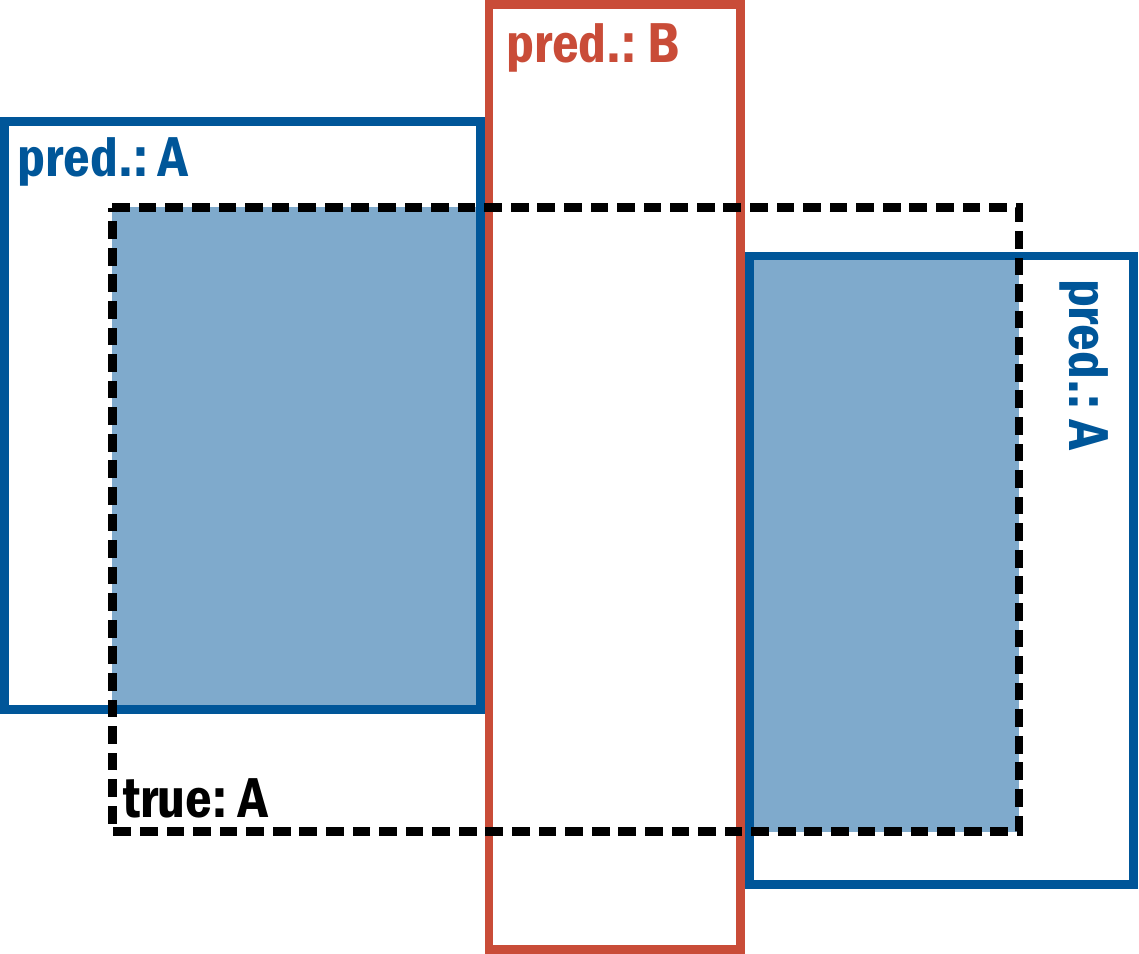}
		\caption{Sketch of a segmentation result.}
		\label{fig:segment-sketch-main}
	\end{subfigure}
	\hfill
	\begin{subfigure}{.35\textwidth}
		\def\svgwidth{\linewidth}
		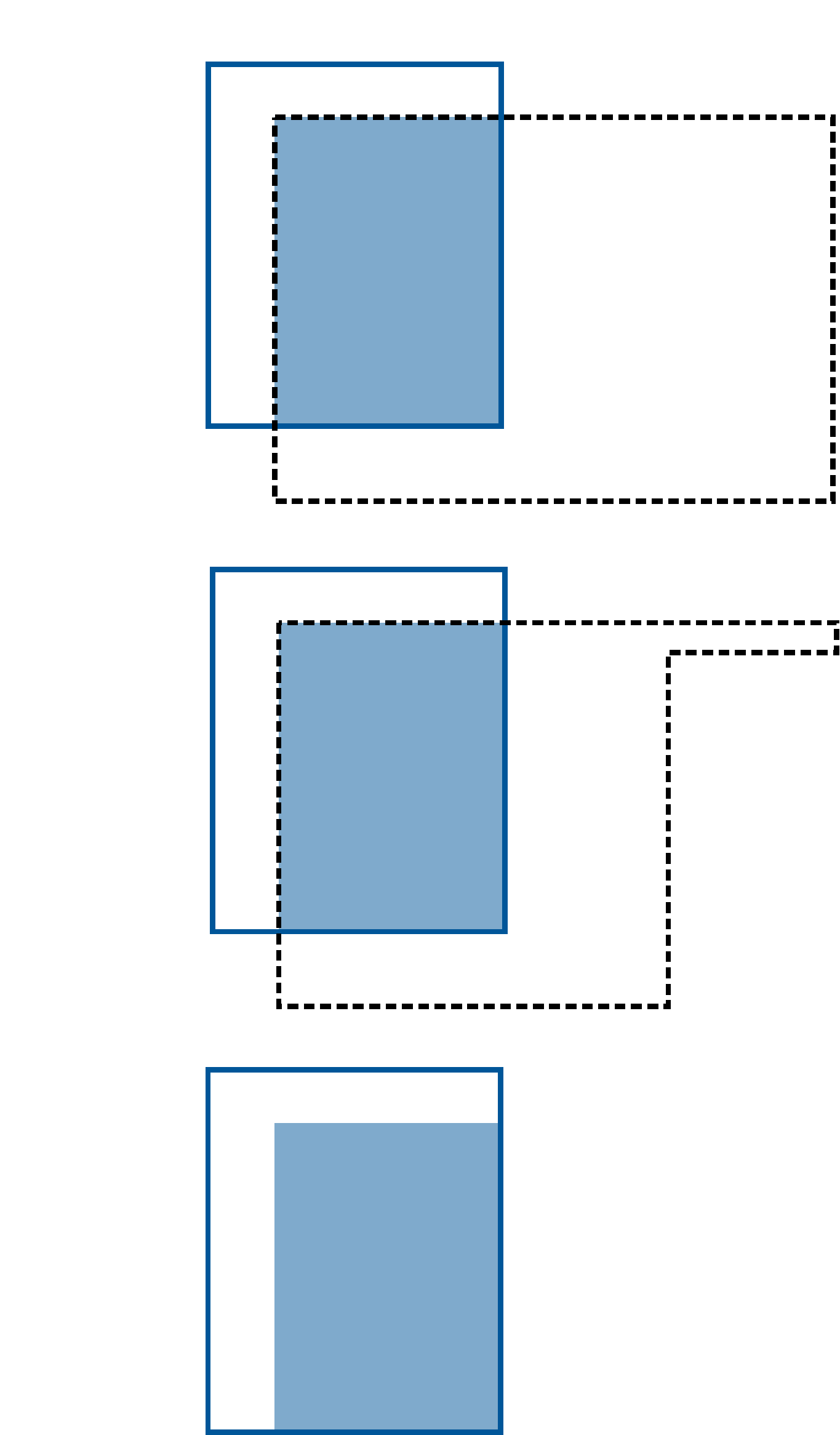
		\caption{Relevant areas for quality metrics.}
		\label{fig:segment-sketch-metrics}
	\end{subfigure}%

	\caption{Sketch of a segmentation result and the quality metrics for one of the segments. 
		(\subref{fig:segment-sketch-main}) A ground truth segment of class \emph{A} (black dashed rectangle) is covered by three predicted segments: two of class \emph{A} (blue), divided by a segment of a different class \emph{B} (red). The correctly segmented area is indicated by the two blue shaded rectangles.
		(\subref{fig:segment-sketch-metrics}) The \IoU of the left-most predicted segment is small, as it is calculated by dividing the blue shaded area by the intersection of the ground truth and the predicted segment, respectively. In contrast, for the $\IoU_{\mathrm{adj.}}$ the area covered by the other segment of class~\emph{A} is disregarded. For the precision, $p$, the correctly predicted area is compared only to the full predicted segment.}
	\label{fig:segment-sketch}		
\end{figure*}

The quality of segments is defined with respect to the ground truth using the measure of intersection over union \citep{Jaccard}. The ground truth segmentation mask is split into a set $\mathcal{K}$ of segments, analogously to the prediction. Predicted segments are then compared to all ground truth segments with a matching class label and a non-trivial intersection, denoted as $\left.\mathcal{K}\right|_{\hat{k}}$. For a predicted segment $\hat{k}\in\hat{\mathcal{K}}$ and the union of the matching and intersecting ground truth segments $K = \bigcup_{k\in \left.\mathcal{K}\right|_{\hat{k}}} k $, the segment-wise intersection over union is defined as 
\[
	\IoU(\hat{k}) = \frac{\left|\hat{k} \cap K\right|}{\left|\hat{k} \cup K\right|}\ .
\] \figurename~\ref{fig:segment-sketch} shows a sketch to clarify the definition of the \IoU and further quality metrics, which are defined and motivated below.

The \IoU penalizes scenarios in which, for example, one ground truth segment is covered by two disjoint predicted segments, which are split by a small, wrongly predicted area. Intuitively, both predicted segments describe a fraction of the ground truth segment well, even though, in the original definition, the \IoU is small.
To address this, the adjusted intersection over union, $\IoU_{\mathrm{adj.}}$, is defined in \cite{metaseg} by restricting the denominator to the union of the predicted segment with the area of the matching ground truth segments which is not covered by other predicted segments of the same class.

In a similar fashion, we assess the quality of predicted segments by their precision, 
\[
	p(\hat{k}) = \frac{\left|\hat{k} \cap K\right|}{\left|\hat{k}\right|}\ ,
\]
i.e., the fraction of pixels in the predicted segment which overlap with a matching ground truth segment. 
For completely wrong predictions, i.e. without overlap of the predicted segment and the ground truth, \mbox{$p=\IoU=\IoU_{\mathrm{adj.}}=0$}. Only for at least partially correct segments, the behavior of the metrics differ and $p \geq \IoU_{\mathrm{adj.}} \geq \IoU$.
By choosing the precision instead of the \IoU, we intentionally neglect to quantify how much of the ground truth segment is covered. For some downstream tasks using the segmentation information of a partial, but precise segment can still be valuable. As an example, a damage detected on a precise but incomplete segment of a car body part is, in many cases, sufficient to provide a correct cost calculation.
	
\FloatBarrier
	
\section{Segment quality classification}

The aforementioned metrics are used to train a segment meta-classifier for a semantic segmentation model for car body parts. The segmentation model is a fully convolutional DNN, distinguishing between 70 car body parts. Segment metrics and ground truth information are collected for about 3000 labeled images, which were used as a validation data set for the training of the segmentation model. An independent set of about 1000 labeled images, which was not used for the training of the segmentation model, provides a test set of segments with ground truth information. 

\begin{figure}
	\centering
	\includegraphics[width=\linewidth]{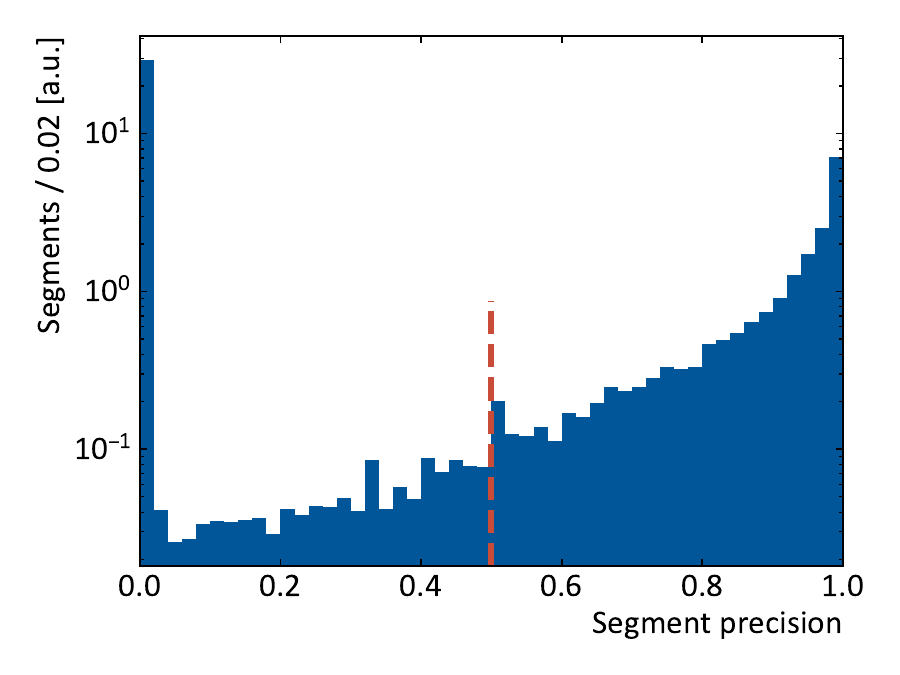}
	\caption{Distribution of the segment-wise precision. Segments with $p>0.5$ are selected as correct predictions. The population of segments at very low precision consists mostly of small, wrongly predicted segments.}
	\label{fig:segments-precision}
\end{figure}

Segments with $p > 0.5$ are labeled as correctly predicted. The threshold, $\tau_p$, is determined from the distribution of the segment precision, c.f. \figurename~\ref{fig:segments-precision}, visual investigation of segments with varying precision and in consideration of downstream tasks. The performance of the meta-classification model does not strongly depend on the chosen precision threshold, as will be detailed below.

Various classification models are trained to predict the binary segment quality, i.e. classify $p>0.5$ versus $p\leq0.5$, and the resulting performance is compared.
Different sets of features are tested, as listed in \tablename~\ref{tab:feature-sets}. 

\begin{table*}
	\centering
	\caption{List of segment-wise features included in the three feature sets: `all', `reduced', and `uncertainty only'.}
	\begin{tabular}{p{.45\textwidth}ccc}
		\toprule
		Features & all & reduced & uncertainty only \\
		\midrule
		Averages of pixel-wise uncertainties & \checkmark & \checkmark & \checkmark \\
		Relative segment size & \checkmark & \checkmark & \\
		Predicted class & \checkmark & \checkmark & \\
		Standard deviation of pixel-wise uncertainties & \checkmark &  &  \\
		Boundary/inner pixel information & \checkmark &  &  \\		
		\bottomrule
	\end{tabular}
	\label{tab:feature-sets}
\end{table*}

Two types of classifiers are tested: a gradient boosted decision tree, based on the XGBoost library \citep{xgboost} as a high performance method~\cite{xgboost-performance}, as well as a linear regression classifier~\citep{scikit-learn}, as a simpler baseline. The XGBoost hyper-parameters are optimized in a grid search employing 5-fold cross validation on the training data set.

\begin{figure}
	\centering
	\includegraphics[width=\linewidth]{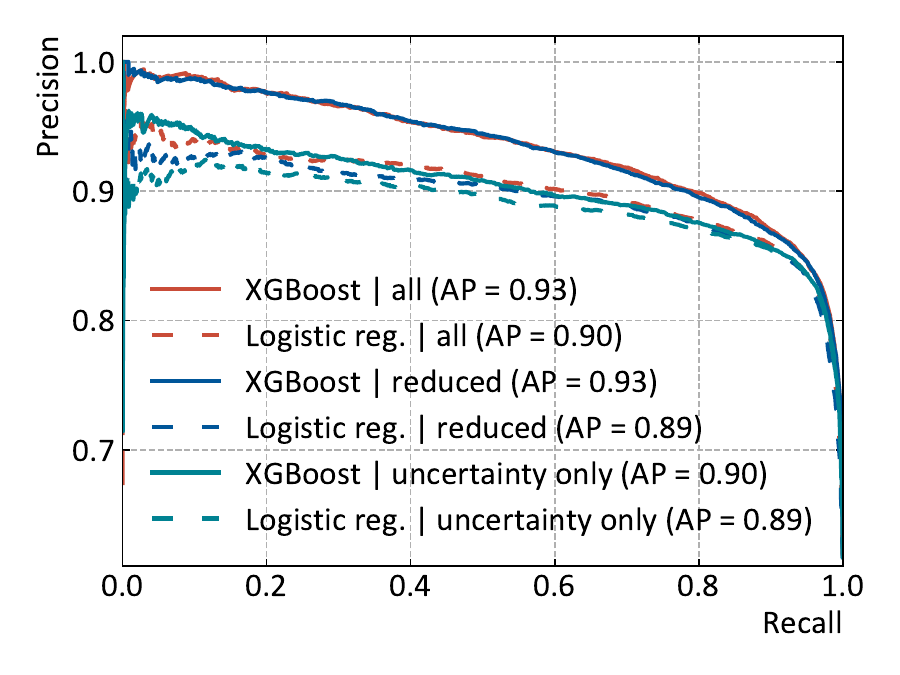}
	\caption{ Precision as a function of the recall obtainable for selecting low-quality segments for all evaluated combinations of classifier types and feature sets. The legend states the average precision $\mathrm{AP}$.}
	\label{fig:precision-recall}
\end{figure}

\tablename~\ref{tab:auroc} lists the area under the receiver operator characteristic curves (AUROC,~\cite{auroc})  obtained for all combinations of classifier types and segment feature sets. The precision-recall curves are displayed in \figurename~\ref{fig:precision-recall}. The XGBoost model trained using all features performs best, achieving an AUROC score of $91.6\%\pm0.2\%$ and an average precision of $93.4\%\pm0.2\%$. Reducing the feature set by excluding the standard deviation of the uncertainty distributions and split of segment features into boundaries and inner areas only entails a minor decrease in performance. 
The achieved AUROC score is $91.5\%\pm0.2\%$ with an average precision of $93.3\%\pm0.2\%$. The results are comparable to the classification results achieved in~\cite{metaseg} for a different model and data set.

The predicted class and the segment size are important for the performance of the XGBoost classifier. Without them, the AUROC score is reduced to $89.0\%\pm0.3\%$ and is on par with the results obtained using the simpler logistic regression of the input features.

\begin{table}
	\centering
	\caption{AUROC scores for all evaluated combinations of classifier types and feature sets, with statistical uncertainties due to the size of the test data set.}
	\label{tab:auroc}
		
	\begin{tabular}{lrr}
		\toprule
		Feature set &  XGBoost &  Log. reg. \\
		\midrule
		all              & 0.916 $\pm$ 0.002 & 0.891 $\pm$ 0.003 \\
		reduced          & 0.915 $\pm$ 0.002 & 0.885 $\pm$ 0.003 \\
		uncertainty only & 0.890 $\pm$ 0.003 & 0.882 $\pm$ 0.003 \\
		\bottomrule
	\end{tabular}

\end{table}

\begin{figure}
	\centering
	\includegraphics[width=\linewidth]{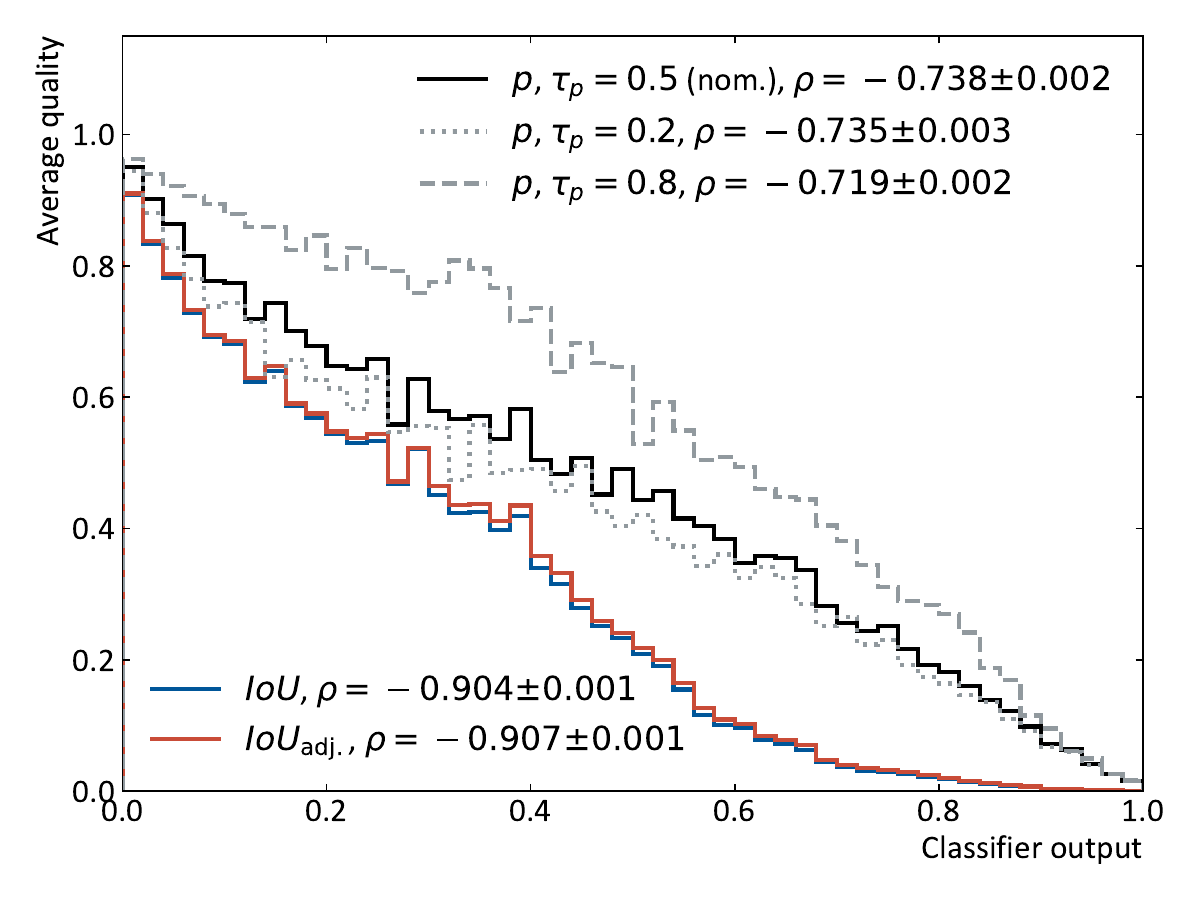}
	\caption{Average segment quality in bins of the meta-classifier output. Shown are $p$ (black), \IoU (blue) and $\IoU_{\mathrm{adj.}}$ (red) for the meta-classifier trained with the nominal precision threshold $\tau_p=0.5$, as well as $p$ for meta-classifiers trained with $\tau_p=0.2$ (gray dotted) and $\tau_p=0.8$ (gray dashed). The legend lists the correlation coefficient $\rho$ for each case.}
	\label{fig:score-vs-quality}
\end{figure}

For further studies, the XGBoost model trained with the reduced feature set is used.
The output of this meta-classification model is scaled to a range of $[0, 1]$ with higher values for segments with a low predicted quality and is used as a measure of the uncertainty for a segment.
As can be seen in \figurename~\ref{fig:score-vs-quality}, the classifier score is strongly correlated with the segment precision ($\rho=0.74$), and the two variants of \IoU ($\rho \geq 0.90$). This correlation prevails even when choosing a different segment precision threshold to define the binary target for meta-classification.

\begin{figure*}[tb]
	\includegraphics[width=\columnwidth]{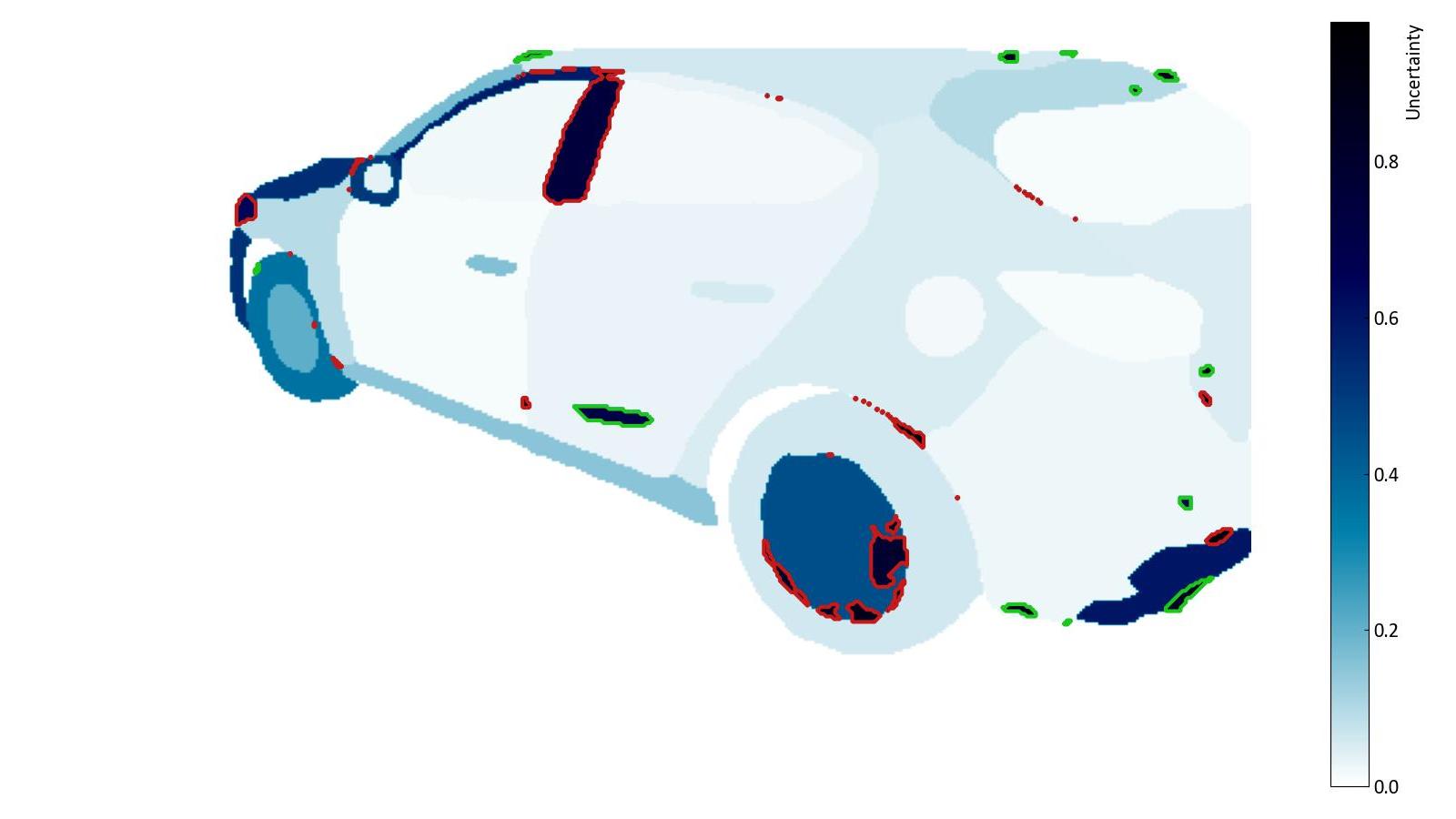}%
	\hfill%
	\includegraphics[width=\columnwidth]{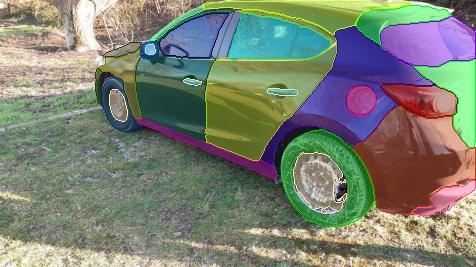}
	\caption{Heat-map of the segment-wise uncertainties (left) and corrected segmentation mask (right) for the example image shown in \figurename~\ref{fig:example-car}. Within the heat-map, the colored contours show segments with an uncertainty above the threshold which are either removed and set to the background class (red), or replaced by the unambiguous surrounding class (green), as decided by the algorithm described in the text.}
	\label{fig:results}
\end{figure*}

\begin{figure*}
	\includegraphics[width=\linewidth]{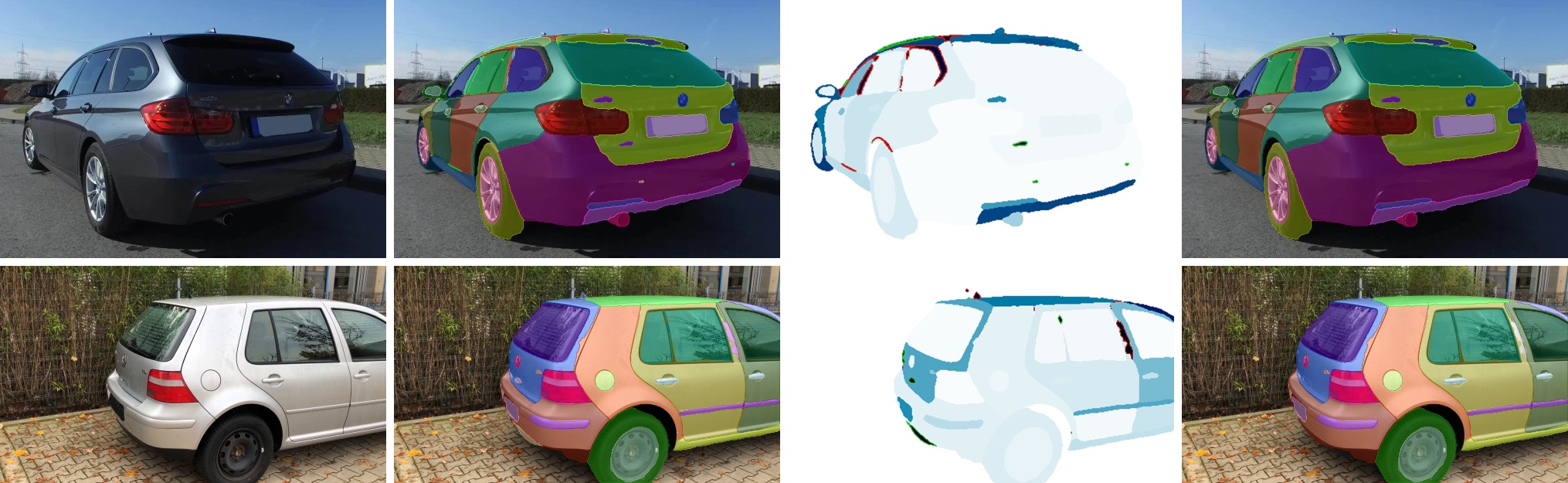}%
	\caption{Additional examples, showing (from left to right) the original image, the segmentation mask, the heat-map of the segment-wise uncertainties and the corrected segmentation mask. Several mistakes, for example on the rear bumper in the upper image and on the trunk in the lower image, are removed.}
	\label{fig:moisaic}
\end{figure*}

The uncertainty measure can be used to remove low-quality segments from the predicted mask. This prevents downstream tasks from including wrong predictions, which can lead to false positive results for car body parts that are not at the predicted location or not even displayed in an image.  
The failure modes of the segmentation model include small, wrongly predicted segments within larger areas of correct predictions. This can be caused by reflections or dirt on the surface of the car. Segments with an uncertainty larger than a specific threshold are removed from the segmentation mask, as detailed in Listing~\ref{alg:correction}. If such a segment is fully enclosed by just one other segment, i.e. if all neighboring pixels have the same predicted class in the original prediction, it is replaced by the enclosing class. Otherwise, the segment is set to the ``background" class, thus preventing downstream tasks from using the pixels for further results. \figurename~\ref{fig:results} shows an example of the segment-wise uncertainties and the corrected segmentation mask for the image shown in \figurename~\ref{fig:example-car}. The wrongly detected air intake segment at the rim is removed, preventing wrong input to subsequent processes. The wrongly predicted molding segment on the door is removed, and replaced by the surrounding door class. 
\figurename~\ref{fig:moisaic} shows additional examples. Comparing the uncertainty map with the segmentation mask and the original image, it can be seen that well segmented parts have low uncertainties, while challenging areas, e.g. due to bad lighting or being in the background of the image, lead to higher segment-wise uncertainties. The mask correction procedure is able to remove many of the erroneously predicted segments.

\begin{algorithm}[t]
	\caption{Segmentation mask correction} \label{alg:correction}
	\begin{algorithmic}
		\ForAll{segments $s$}
			\If{$\Call{Uncertainty}{s} > \tau$}
				\State | {\footnotesize Collect neighbor segments, for example using} 
				\State | {\footnotesize a dilation operation on the pixel mask}
				\State $n \gets \Call{NeighborSegments}{s}$ 
				\If{$\Call{len}{n} = 1$}
					\State $\Call{class}{s} \gets \Call{class}{n_0}$
				\Else
					\State $\Call{class}{s} \gets \textsc{background}$
				\EndIf
			\EndIf
		\EndFor
	\end{algorithmic}
\end{algorithm}

The segment-wise uncertainty map provides comprehensive and easy-to-use information about the reliability of each segment for further applications. For example, if damages are found only on segments with a low uncertainty, the claims handling process can be automated with high confidence in the end result. Individual high uncertainty segments can be removed from the segmentation mask, in order to improve the quality of the result.

The quality of a segmentation mask for an image can be characterized by the mean (i.e., class averaged) \IoU. Given the sets of predicted classes, $\hat{\mathcal{C}}$, and of the classes in the ground truth labels, $\mathcal{C}$, for an image, this metric is defined as 
\[m\IoU = \frac{1}{\left|\hat{\mathcal{C}}\cup\mathcal{C}\right|} \sum_{c\in\hat{\mathcal{C}}\cup\mathcal{C}} \frac{tp_c}{tp_c + fp_c + fn_c} \ ,\]
where $tp_c$, $fp_c$, and $fn_c$ are the numbers of true positive, false positive and false negative predicted pixels of class $c$, respectively.
Notably, any class which is neither in the prediction nor in the labels does not affect the $m\IoU$, while classes which are in the predicted segments but not in the ground truth labels (and vice versa) reduce the $m\IoU$ of a segmentation mask.

\begin{figure}
	\centering
	\includegraphics[width=\linewidth]{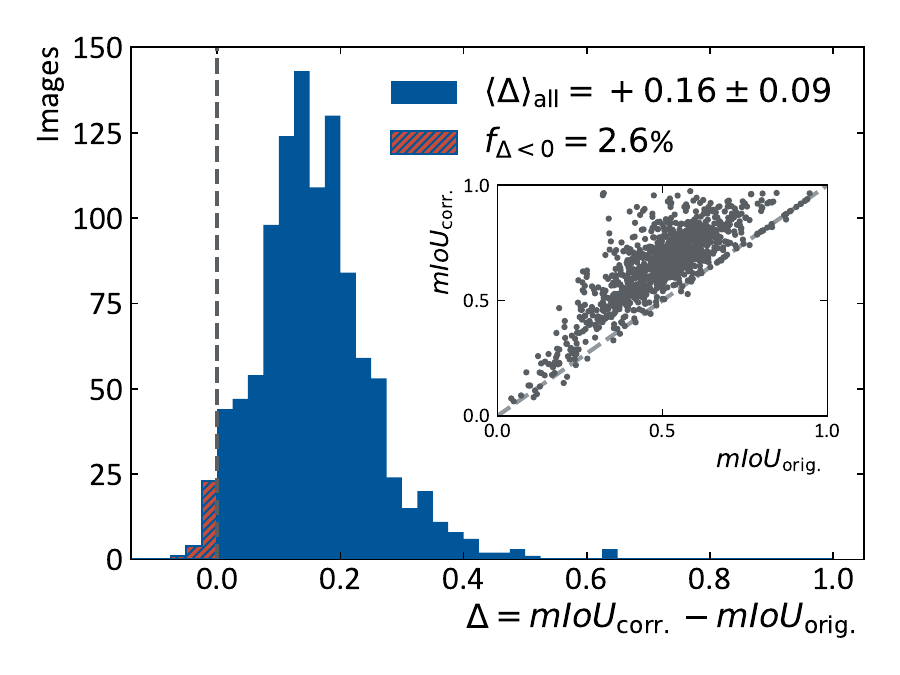}
	\caption{Distribution of the $m\IoU$ difference, $\Delta$, between the corrected and the original mask. The red, hatched area marks entries with $\Delta < 0$, indicating a quality degradation, occurring only for $f_{\Delta < 0} = 2.6\%$ of the images. The inset shows a scatter plot of the $m\IoU$ of the corrected prediction in dependence on the $m\IoU$ of the original mask.}
	\label{fig:delta-miou}
\end{figure}

The $m\IoU$ is computed image by image for the original segmentation mask, as well as for the corrected mask, to quantify the impact of removing segments with a high uncertainty. \figurename~\ref{fig:delta-miou} shows the distribution of the difference between these two values. On average over all images, the $m\IoU$ is improved by $\overline{\Delta m\IoU} = 0.16$, corresponding to an increase of the average $m\IoU$ from $0.50$ to $0.66$. The standard deviation of the distribution of $\Delta m\IoU$ is $0.09$ on the test set.
For $>97\%$ of the images in the test set an improvement of the result is observed. In the rare cases that the correction procedure results in a $m\IoU$ decrease, usually small, irregularly formed but precise segments within the larger area of a misidentified car body part are removed.
 \figurename~\ref{fig:delta-miou} also shows the $m\IoU$ values for corrected masks in dependence on the uncorrected result. The method yields improvements over a large range of $m\IoU$.
 
In order to study the robustness of the correction procedure, images are grouped into different categories. Different image perspectives bring different challenges to the model: images showing the full car have smaller relative segment sizes, while zoom images can lack helpful context. The exposure of the image could have an impact on the procedure, as under- or over-exposed areas effectively hide information. Lastly, the image resolution is an important factor for the overall image quality. Table~\ref{tab:robustness} lists the average improvement $\Delta m\IoU$ due to the correction procedure for images in different categories. The individual results agree well with the overall average, showing that the method is robust under the tested effects. 

A major factor of the improvement is the removal of small segments, in turn leading to a wrongly predicted class being removed from the mask entirely. Even though only a small fraction of pixels in the image is affected, the effect on the $m\IoU$ is significant because every class has the same weight.
The number of wrongly predicted classes per image is reduced from 6.3 to 1.4, on average, with standard deviations of 4.0 and 1.5, respectively. At the same time, the a small decrease in the number of correctly predicted classes is observed as well, reducing the number from 11.2 to 10.6, with standard deviations of 7.3 and 6.9.
Crucially, this reduction prevents false positive detections in downstream tasks.

\begin{table}[tb]
	\centering
	\caption{Average $\Delta m\IoU$ for images in different categories of image perspective, exposure and resolution.}
	\begin{tabular}{ccc}
		\toprule
		& \textbf{Perspective} & \\
		Full car & & Zoom \\
		\midrule
		$0.16\pm0.11$ & & $0.17 \pm 0.07$ \\
		\bottomrule
		\addlinespace[1ex]
		& \textbf{Exposure} & \\
		Underexposed & Balanced exposure & Overexposed \\
		\midrule
		$0.14 \pm 0.09$ & $0.17\pm0.10$ & $0.17 \pm 0.10$ \\		
		\bottomrule
		\addlinespace[1ex]
		& \textbf{Resolution} & \\
		$<1$\,MP & $1-4$\,MP & $>4$\,MP \\
		\midrule
		$0.17 \pm 0.11$ & $0.17\pm0.10$ & $0.18 \pm 0.07$ \\		
		\bottomrule		
	\end{tabular}
	\label{tab:robustness}
\end{table}

\section{Conclusion}
In this work, the development and application of a meta-classification model is presented, which is used to assess the quality of the output of a semantic segmentation model for car body parts.
Pixel-wise uncertainties are derived from the softmax probabilities and gradients, and are combined to segment-wise features. A gradient boosted decision tree classifier based on the average uncertainty features per segment has been trained to distinguish between precise and imprecise segments. The resulting meta-model achieves an AUROC score of $0.915\pm0.002$. The outputs of this classifier provide a comprehensive uncertainty measure for each segment.

In a production setting, the meta-classification model runs as a post-processing step after evaluating the car body part segmentation model.  The resulting uncertainty scores are then used to 
remove low-quality segments from the predictions. This removal prevents false positive detections in downstream tasks and improves the segmentation mask quality for this use-case by $\overline{\Delta m\IoU} = 0.16$. 

The proposed method can improve the reliability of a segmentation model output. In the context of motor claims handling, this has been proven to be a valuable tool for the automation of damage assessment tasks.

\section*{Statements and Declarations}
The authors declare that they have no known competing financial interests or personal
relationships that could have appeared to influence the work reported in this paper.

\section*{Acknowledgments}
We thank Hanno Gottschalk, Matthias Rottmann, Svenja Uhlemeyer and the IZMD at Bergische Universität Wuppertal for helpful discussions and useful advice. 

\bibliography{references} 	
	
\end{document}

%% file: figures/text/segment_sketch_metrics.pdf_tex
%% Creator: Inkscape 1.2.2 (732a01da63, 2022-12-09), www.inkscape.org
%% Accompanies image file 'segment_sketch_metrics.pdf' (pdf, eps, ps)
\begingroup%
  \makeatletter%
  \providecommand\color[2][]{%
    \errmessage{(Inkscape) Color is used for the text in Inkscape, but the package 'color.sty' is not loaded}%
    \renewcommand\color[2][]{}%
  }%
  \providecommand\transparent[1]{%
    \errmessage{(Inkscape) Transparency is used (non-zero) for the text in Inkscape, but the package 'transparent.sty' is not loaded}%
    \renewcommand\transparent[1]{}%
  }%
  \providecommand\rotatebox[2]{#2}%
  \newcommand*\fsize{\dimexpr\f@size pt\relax}%
  \newcommand*\lineheight[1]{\fontsize{\fsize}{#1\fsize}\selectfont}%
  \ifx\svgwidth\undefined%
    \setlength{\unitlength}{654.82577106bp}%
    \ifx\svgscale\undefined%
      \relax%
    \else%
      \setlength{\unitlength}{\unitlength * \real{\svgscale}}%
    \fi%
  \else%
    \setlength{\unitlength}{\svgwidth}%
  \fi%
  \global\let\svgwidth\undefined%
  \global\let\svgscale\undefined%
  \makeatother%
  \begin{picture}(1,1.70904714)%
    \lineheight{1}%
    \setlength\tabcolsep{0pt}%
    \put(0,0){\includegraphics[width=\unitlength,page=1]{figures/text/segment_sketch_metrics.pdf}}%
    \put(0.96767263,1.33257198){\color[rgb]{0,0,0}\makebox(0,0)[rt]{\lineheight{1}\smash{\footnotesize\begin{tabular}[t]{r}false\\negative\\area\end{tabular}}}}%
    \put(0.33590733,1.43393706){\color[rgb]{0,0,0}\makebox(0,0)[lt]{\lineheight{1}\smash{\footnotesize\begin{tabular}[t]{l} true\\positive\\area\end{tabular}}}}%
    \put(0.43843619,1.66231715){\color[rgb]{0,0,0}\makebox(0,0)[lt]{\lineheight{1}\smash{\footnotesize\begin{tabular}[t]{l}false positive area\end{tabular}}}}%
    \put(0.09023732,1.33711749){\color[rgb]{0,0,0}\makebox(0,0)[lt]{\lineheight{1.25}\smash{\IoU}}}%
    \put(-0.00360902,0.82074808){\color[rgb]{0,0,0}\makebox(0,0)[lt]{\lineheight{1.25}\smash{$\IoU_{\mathrm{adj.}}$}}}%
    \put(0.15949427,0.22348475){\color[rgb]{0,0,0}\makebox(0,0)[lt]{\lineheight{1.25}\smash{$p$}}}%
    \put(0,0){\includegraphics[width=\unitlength,page=2]{figures/text/segment_sketch_metrics.pdf}}%
  \end{picture}%
\endgroup%

%% file: paper.bbl

\begin{thebibliography}{26}
\ifx \bisbn   \undefined \def \bisbn  #1{ISBN #1}\fi
\ifx \binits  \undefined \def \binits#1{#1}\fi
\ifx \bauthor  \undefined \def \bauthor#1{#1}\fi
\ifx \batitle  \undefined \def \batitle#1{#1}\fi
\ifx \bjtitle  \undefined \def \bjtitle#1{#1}\fi
\ifx \bvolume  \undefined \def \bvolume#1{\textbf{#1}}\fi
\ifx \byear  \undefined \def \byear#1{#1}\fi
\ifx \bissue  \undefined \def \bissue#1{#1}\fi
\ifx \bfpage  \undefined \def \bfpage#1{#1}\fi
\ifx \blpage  \undefined \def \blpage #1{#1}\fi
\ifx \burl  \undefined \def \burl#1{\textsf{#1}}\fi
\ifx \doiurl  \undefined \def \doiurl#1{\url{https://doi.org/#1}}\fi
\ifx \betal  \undefined \def \betal{\textit{et al.}}\fi
\ifx \binstitute  \undefined \def \binstitute#1{#1}\fi
\ifx \binstitutionaled  \undefined \def \binstitutionaled#1{#1}\fi
\ifx \bctitle  \undefined \def \bctitle#1{#1}\fi
\ifx \beditor  \undefined \def \beditor#1{#1}\fi
\ifx \bpublisher  \undefined \def \bpublisher#1{#1}\fi
\ifx \bbtitle  \undefined \def \bbtitle#1{#1}\fi
\ifx \bedition  \undefined \def \bedition#1{#1}\fi
\ifx \bseriesno  \undefined \def \bseriesno#1{#1}\fi
\ifx \blocation  \undefined \def \blocation#1{#1}\fi
\ifx \bsertitle  \undefined \def \bsertitle#1{#1}\fi
\ifx \bsnm \undefined \def \bsnm#1{#1}\fi
\ifx \bsuffix \undefined \def \bsuffix#1{#1}\fi
\ifx \bparticle \undefined \def \bparticle#1{#1}\fi
\ifx \barticle \undefined \def \barticle#1{#1}\fi
\bibcommenthead
\ifx \bconfdate \undefined \def \bconfdate #1{#1}\fi
\ifx \botherref \undefined \def \botherref #1{#1}\fi
\ifx \url \undefined \def \url#1{\textsf{#1}}\fi
\ifx \bchapter \undefined \def \bchapter#1{#1}\fi
\ifx \bbook \undefined \def \bbook#1{#1}\fi
\ifx \bcomment \undefined \def \bcomment#1{#1}\fi
\ifx \oauthor \undefined \def \oauthor#1{#1}\fi
\ifx \citeauthoryear \undefined \def \citeauthoryear#1{#1}\fi
\ifx \endbibitem  \undefined \def \endbibitem {}\fi
\ifx \bconflocation  \undefined \def \bconflocation#1{#1}\fi
\ifx \arxivurl  \undefined \def \arxivurl#1{\textsf{#1}}\fi
\csname PreBibitemsHook\endcsname

%%% 1
\bibitem[\protect\citeauthoryear{Shotton et~al.}{2006}]{semantic-segmentation}
\begin{bchapter}
\bauthor{\bsnm{Shotton}, \binits{J.}},
\bauthor{\bsnm{Winn}, \binits{J.}},
\bauthor{\bsnm{Rother}, \binits{C.}},
\bauthor{\bsnm{Criminisi}, \binits{A.}}:
\bctitle{{TextonBoost}: Joint appearance, shape and context modeling for
  multi-class object recognition and segmentation}.
In: \beditor{\bsnm{Leonardis}, \binits{A.}},
\beditor{\bsnm{Bischof}, \binits{H.}},
\beditor{\bsnm{Pinz}, \binits{A.}} (eds.)
\bbtitle{Computer Vision -- ECCV 2006},
pp. \bfpage{1}--\blpage{15}.
\bpublisher{Springer},
\blocation{Berlin, Heidelberg}
(\byear{2006}).
\doiurl{10.1007/11744023_1}
\end{bchapter}
\endbibitem

%%% 2
\bibitem[\protect\citeauthoryear{Wang et~al.}{2021}]{segmentation-hrnet}
\begin{barticle}
\bauthor{\bsnm{Wang}, \binits{J.}},
\bauthor{\bsnm{Sun}, \binits{K.}},
\bauthor{\bsnm{Cheng}, \binits{T.}},
\bauthor{\bsnm{Jiang}, \binits{B.}},
\bauthor{\bsnm{Deng}, \binits{C.}},
\bauthor{\bsnm{Zhao}, \binits{Y.}},
\bauthor{\bsnm{Liu}, \binits{D.}},
\bauthor{\bsnm{Mu}, \binits{Y.}},
\bauthor{\bsnm{Tan}, \binits{M.}},
\bauthor{\bsnm{Wang}, \binits{X.}},
\bauthor{\bsnm{Liu}, \binits{W.}},
\bauthor{\bsnm{Xiao}, \binits{B.}}:
\batitle{Deep high-resolution representation learning for visual recognition}.
\bjtitle{IEEE Transactions on Pattern Analysis and Machine Intelligence}
\bvolume{43}(\bissue{10}),
\bfpage{3349}--\blpage{3364}
(\byear{2021})
\doiurl{10.1109/TPAMI.2020.2983686}
\end{barticle}
\endbibitem

%%% 3
\bibitem[\protect\citeauthoryear{Yu et~al.}{2023}]{segmentation-review}
\begin{botherref}
\oauthor{\bsnm{Yu}, \binits{Y.}},
\oauthor{\bsnm{Wang}, \binits{C.}},
\oauthor{\bsnm{Fu}, \binits{Q.}},
\oauthor{\bsnm{Kou}, \binits{R.}},
\oauthor{\bsnm{Huang}, \binits{F.}},
\oauthor{\bsnm{Yang}, \binits{B.}},
\oauthor{\bsnm{Yang}, \binits{T.}},
\oauthor{\bsnm{Gao}, \binits{M.}}:
Techniques and challenges of image segmentation: A review.
Electronics
\textbf{12}(5)
(2023)
\doiurl{10.3390/electronics12051199}
\end{botherref}
\endbibitem

%%% 4
\bibitem[\protect\citeauthoryear{de~Jorge et~al.}{2023}]{reliabilitytrack}
\begin{bchapter}
\bauthor{\bsnm{Jorge}, \binits{P.}},
\bauthor{\bsnm{Volpi}, \binits{R.}},
\bauthor{\bsnm{Torr}, \binits{P.H.S.}},
\bauthor{\bsnm{Rogez}, \binits{G.}}:
\bctitle{Reliability in semantic segmentation: Are we on the right track?}
In: \bbtitle{2023 IEEE/CVF Conference on Computer Vision and Pattern
  Recognition (CVPR)},
pp. \bfpage{7173}--\blpage{7182}
(\byear{2023}).
\doiurl{10.1109/CVPR52729.2023.00693}
\end{bchapter}
\endbibitem

%%% 5
\bibitem[\protect\citeauthoryear{Mehrtash et~al.}{2020}]{score-confidence}
\begin{barticle}
\bauthor{\bsnm{Mehrtash}, \binits{A.}},
\bauthor{\bsnm{Wells}, \binits{W.M.}},
\bauthor{\bsnm{Tempany}, \binits{C.M.}},
\bauthor{\bsnm{Abolmaesumi}, \binits{P.}},
\bauthor{\bsnm{Kapur}, \binits{T.}}:
\batitle{Confidence calibration and predictive uncertainty estimation for deep
  medical image segmentation}.
\bjtitle{{IEEE} Transactions on Medical Imaging}
\bvolume{39}(\bissue{12}),
\bfpage{3868}--\blpage{3878}
(\byear{2020})
\doiurl{10.1109/tmi.2020.3006437}
{\href{https://arxiv.org/abs/1911.13273}{{1911.13273}}}
\end{barticle}
\endbibitem

%%% 6
\bibitem[\protect\citeauthoryear{Hein et~al.}{2019}]{relu-overconfidence}
\begin{bchapter}
\bauthor{\bsnm{Hein}, \binits{M.}},
\bauthor{\bsnm{Andriushchenko}, \binits{M.}},
\bauthor{\bsnm{Bitterwolf}, \binits{J.}}:
\bctitle{Why {ReLU} networks yield high-confidence predictions far away from
  the training data and how to mitigate the problem}.
In: \bbtitle{2019 IEEE/CVF Conference on Computer Vision and Pattern
  Recognition (CVPR)},
\bconflocation{Long Beach, CA, USA},
pp. \bfpage{41}--\blpage{50}
(\byear{2019}).
\doiurl{10.1109/CVPR.2019.00013}
\end{bchapter}
\endbibitem

%%% 7
\bibitem[\protect\citeauthoryear{Guo et~al.}{2017}]{confidence-calibration}
\begin{bchapter}
\bauthor{\bsnm{Guo}, \binits{C.}},
\bauthor{\bsnm{Pleiss}, \binits{G.}},
\bauthor{\bsnm{Sun}, \binits{Y.}},
\bauthor{\bsnm{Weinberger}, \binits{K.Q.}}:
\bctitle{On calibration of modern neural networks}.
In: \bbtitle{Proceedings of the 34th International Conference on Machine
  Learning - Volume 70}.
\bsertitle{ICML'17},
pp. \bfpage{1321}--\blpage{1330}.
\bpublisher{JMLR.org},
\blocation{Sydney, NSW, Australia}
(\byear{2017})
\end{bchapter}
\endbibitem

%%% 8
\bibitem[\protect\citeauthoryear{Abdar et~al.}{2021}]{DeepLearningUQReview}
\begin{barticle}
\bauthor{\bsnm{Abdar}, \binits{M.}},
\bauthor{\bsnm{Pourpanah}, \binits{F.}},
\bauthor{\bsnm{Hussain}, \binits{S.}},
\bauthor{\bsnm{Rezazadegan}, \binits{D.}},
\bauthor{\bsnm{Liu}, \binits{L.}},
\bauthor{\bsnm{Ghavamzadeh}, \binits{M.}},
\bauthor{\bsnm{Fieguth}, \binits{P.}},
\bauthor{\bsnm{Cao}, \binits{X.}},
\bauthor{\bsnm{Khosravi}, \binits{A.}},
\bauthor{\bsnm{Acharya}, \binits{U.R.}},
\bauthor{\bsnm{Makarenkov}, \binits{V.}},
\bauthor{\bsnm{Nahavandi}, \binits{S.}}:
\batitle{A review of uncertainty quantification in deep learning: Techniques,
  applications and challenges}.
\bjtitle{Information Fusion}
\bvolume{76},
\bfpage{243}--\blpage{297}
(\byear{2021})
\doiurl{10.1016/j.inffus.2021.05.008}
\end{barticle}
\endbibitem

%%% 9
\bibitem[\protect\citeauthoryear{Dechesne et~al.}{2021}]{BayesianUNet}
\begin{botherref}
\oauthor{\bsnm{Dechesne}, \binits{C.}},
\oauthor{\bsnm{Lassalle}, \binits{P.}},
\oauthor{\bsnm{Lefèvre}, \binits{S.}}:
{Bayesian U-Net: Estimating Uncertainty in Semantic Segmentation of Earth
  Observation Images}.
Remote Sensing
\textbf{13}(19)
(2021)
\doiurl{10.3390/rs13193836}
\end{botherref}
\endbibitem

%%% 10
\bibitem[\protect\citeauthoryear{Kendall et~al.}{2017}]{bayesian-segnet}
\begin{bchapter}
\bauthor{\bsnm{Kendall}, \binits{A.}},
\bauthor{\bsnm{Badrinarayananm}, \binits{V.}},
\bauthor{\bsnm{Cipolla}, \binits{R.}}:
\bctitle{{Bayesian SegNet: Model Uncertainty in Deep Convolutional
  Encoder-Decoder Architectures for Scene Understanding}}.
In: \beditor{\bsnm{Kim}, \binits{T.-K.}},
\beditor{\bsnm{Zafeiriou}, \binits{S.}},
\beditor{\bsnm{Brostow}, \binits{G.}},
\beditor{\bsnm{Mikolajczyk}, \binits{K.}} (eds.)
\bbtitle{Proceedings of the British Machine Vision Conference (BMVC)},
pp. \bfpage{1}--\blpage{12}.
\bpublisher{BMVA Press},
\blocation{London, United Kingdom}
(\byear{2017}).
\doiurl{10.5244/C.31.57}
\end{bchapter}
\endbibitem

%%% 11
\bibitem[\protect\citeauthoryear{Wang
  et~al.}{2019}]{SegmentationDropoutAugmentation}
\begin{barticle}
\bauthor{\bsnm{Wang}, \binits{G.}},
\bauthor{\bsnm{Li}, \binits{W.}},
\bauthor{\bsnm{Aertsen}, \binits{M.}},
\bauthor{\bsnm{Deprest}, \binits{J.}},
\bauthor{\bsnm{Ourselin}, \binits{S.}},
\bauthor{\bsnm{Vercauteren}, \binits{T.}}:
\batitle{Aleatoric uncertainty estimation with test-time augmentation for
  medical image segmentation with convolutional neural networks}.
\bjtitle{Neurocomputing}
\bvolume{338},
\bfpage{34}--\blpage{45}
(\byear{2019})
\doiurl{10.1016/j.neucom.2019.01.103}
\end{barticle}
\endbibitem

%%% 12
\bibitem[\protect\citeauthoryear{Blundell et~al.}{2015}]{BBB}
\begin{bchapter}
\bauthor{\bsnm{Blundell}, \binits{C.}},
\bauthor{\bsnm{Cornebise}, \binits{J.}},
\bauthor{\bsnm{Kavukcuoglu}, \binits{K.}},
\bauthor{\bsnm{Wierstra}, \binits{D.}}:
\bctitle{Weight uncertainty in neural network}.
In: \beditor{\bsnm{Bach}, \binits{F.}},
\beditor{\bsnm{Blei}, \binits{D.}} (eds.)
\bbtitle{Proceedings of the 32nd International Conference on Machine Learning}.
\bsertitle{Proceedings of Machine Learning Research},
vol. \bseriesno{37},
pp. \bfpage{1613}--\blpage{1622}.
\bpublisher{PMLR},
\blocation{Lille, France}
(\byear{2015})
\end{bchapter}
\endbibitem

%%% 13
\bibitem[\protect\citeauthoryear{Lee et~al.}{2020}]{ensemble}
\begin{botherref}
\oauthor{\bsnm{Lee}, \binits{H.J.}},
\oauthor{\bsnm{Kim}, \binits{S.T.}},
\oauthor{\bsnm{Lee}, \binits{H.}},
\oauthor{\bsnm{Navab}, \binits{N.}},
\oauthor{\bsnm{Ro}, \binits{Y.M.}}:
Efficient Ensemble Model Generation for Uncertainty Estimation with Bayesian
  Approximation in Segmentation
(2020)
\end{botherref}
\endbibitem

%%% 14
\bibitem[\protect\citeauthoryear{Ng
  et~al.}{2023}]{SegmentationUncertaintyReview}
\begin{barticle}
\bauthor{\bsnm{Ng}, \binits{M.}},
\bauthor{\bsnm{Guo}, \binits{F.}},
\bauthor{\bsnm{Biswas}, \binits{L.}},
\bauthor{\bsnm{Petersen}, \binits{S.E.}},
\bauthor{\bsnm{Piechnik}, \binits{S.K.}},
\bauthor{\bsnm{Neubauer}, \binits{S.}},
\bauthor{\bsnm{Wright}, \binits{G.}}:
\batitle{{Estimating Uncertainty in Neural Networks for Cardiac MRI
  Segmentation: A Benchmark Study}}.
\bjtitle{IEEE Transactions on Biomedical Engineering}
\bvolume{70}(\bissue{6}),
\bfpage{1955}--\blpage{1966}
(\byear{2023})
\doiurl{10.1109/TBME.2022.3232730}
\end{barticle}
\endbibitem

%%% 15
\bibitem[\protect\citeauthoryear{Landgraf et~al.}{2023}]{DUDESdistillation}
\begin{botherref}
\oauthor{\bsnm{Landgraf}, \binits{S.}},
\oauthor{\bsnm{Wursthorn}, \binits{K.}},
\oauthor{\bsnm{Hillemann}, \binits{M.}},
\oauthor{\bsnm{Ulrich}, \binits{M.}}:
{DUDES: Deep Uncertainty Distillation using Ensembles for Semantic
  Segmentation}
(2023).
\doiurl{10.48550/arXiv.2303.09843}
\end{botherref}
\endbibitem

%%% 16
\bibitem[\protect\citeauthoryear{Holder and Shafique}{2021}]{distillation}
\begin{bchapter}
\bauthor{\bsnm{Holder}, \binits{C.J.}},
\bauthor{\bsnm{Shafique}, \binits{M.}}:
\bctitle{Efficient uncertainty estimation in semantic segmentation via
  distillation}.
In: \bbtitle{2021 IEEE/CVF International Conference on Computer Vision
  Workshops (ICCVW)},
pp. \bfpage{3080}--\blpage{3087}
(\byear{2021}).
\doiurl{10.1109/ICCVW54120.2021.00343}
\end{bchapter}
\endbibitem

%%% 17
\bibitem[\protect\citeauthoryear{Lin and Hauptmann}{2003}]{metaclassification}
\begin{bchapter}
\bauthor{\bsnm{Lin}, \binits{W.-H.}},
\bauthor{\bsnm{Hauptmann}, \binits{A.}}:
\bctitle{Meta-classification: Combining multimodal classifiers}.
In: \beditor{\bsnm{Za{\"i}ane}, \binits{O.R.}},
\beditor{\bsnm{Simoff}, \binits{S.J.}},
\beditor{\bsnm{Djeraba}, \binits{C.}} (eds.)
\bbtitle{Mining Multimedia and Complex Data},
pp. \bfpage{217}--\blpage{231}.
\bpublisher{Springer},
\blocation{Berlin, Heidelberg}
(\byear{2003}).
\doiurl{10.1007/978-3-540-39666-6_14}
\end{bchapter}
\endbibitem

%%% 18
\bibitem[\protect\citeauthoryear{Rottmann et~al.}{2020}]{metaseg}
\begin{bchapter}
\bauthor{\bsnm{Rottmann}, \binits{M.}},
\bauthor{\bsnm{Colling}, \binits{P.}},
\bauthor{\bsnm{Paul~Hack}, \binits{T.}},
\bauthor{\bsnm{Chan}, \binits{R.}},
\bauthor{\bsnm{Huger}, \binits{F.}},
\bauthor{\bsnm{Schlicht}, \binits{P.}},
\bauthor{\bsnm{Gottschalk}, \binits{H.}}:
\bctitle{Prediction error meta classification in semantic segmentation:
  Detection via aggregated dispersion measures of softmax probabilities}.
In: \bbtitle{2020 International Joint Conference on Neural Networks ({IJCNN})}.
\bpublisher{IEEE},
\blocation{Glasgow, United Kingdom}
(\byear{2020}).
\doiurl{10.1109/IJCNN48605.2020.9206659}
\end{bchapter}
\endbibitem

%%% 19
\bibitem[\protect\citeauthoryear{Hendrycks and Gimpel}{2017}]{softmax-baseline}
\begin{bchapter}
\bauthor{\bsnm{Hendrycks}, \binits{D.}},
\bauthor{\bsnm{Gimpel}, \binits{K.}}:
\bctitle{A baseline for detecting misclassified and out-of-distribution
  examples in neural networks}.
In: \bbtitle{5th International Conference on Learning Representations, {ICLR}
  2017}.
\bpublisher{OpenReview.net},
\blocation{Toulon, France}
(\byear{2017}).
\burl{https://openreview.net/forum?id=Hkg4TI9xl}
\end{bchapter}
\endbibitem

%%% 20
\bibitem[\protect\citeauthoryear{Maag and
  Riedlinger}{2023}]{gradient-uncertainty}
\begin{botherref}
\oauthor{\bsnm{Maag}, \binits{K.}},
\oauthor{\bsnm{Riedlinger}, \binits{T.}}:
Pixel-wise Gradient Uncertainty for Convolutional Neural Networks applied to
  Out-of-Distribution Segmentation
(2023).
\doiurl{10.48550/arXiv.2303.06920}
\end{botherref}
\endbibitem

%%% 21
\bibitem[\protect\citeauthoryear{Grcić et~al.}{2023}]{MaskLevelAdvantages}
\begin{botherref}
\oauthor{\bsnm{Grcić}, \binits{M.}},
\oauthor{\bsnm{Šarić}, \binits{J.}},
\oauthor{\bsnm{Šegvić}, \binits{S.}}:
On Advantages of Mask-level Recognition for Outlier-aware Segmentation
(2023).
\doiurl{10.48550/arXiv.2301.03407}
\end{botherref}
\endbibitem

%%% 22
\bibitem[\protect\citeauthoryear{Jaccard}{1912}]{Jaccard}
\begin{barticle}
\bauthor{\bsnm{Jaccard}, \binits{P.}}:
\batitle{The distribution of the flora in the alpine zone}.
\bjtitle{The New Phytologist}
\bvolume{11}(\bissue{2}),
\bfpage{37}--\blpage{50}
(\byear{1912})
\doiurl{10.1111/j.1469-8137.1912.tb05611.x}
\end{barticle}
\endbibitem

%%% 23
\bibitem[\protect\citeauthoryear{Chen and Guestrin}{2016}]{xgboost}
\begin{bchapter}
\bauthor{\bsnm{Chen}, \binits{T.}},
\bauthor{\bsnm{Guestrin}, \binits{C.}}:
\bctitle{{XGBoost}}.
In: \bbtitle{Proceedings of the 22nd {ACM} {SIGKDD} International Conference on
  Knowledge Discovery and Data Mining}.
\bpublisher{ACM},
\blocation{New York, NY, USA}
(\byear{2016}).
\doiurl{10.1145/2939672.2939785}
\end{bchapter}
\endbibitem

%%% 24
\bibitem[\protect\citeauthoryear{Shwartz-Ziv and
  Armon}{2022}]{xgboost-performance}
\begin{barticle}
\bauthor{\bsnm{Shwartz-Ziv}, \binits{R.}},
\bauthor{\bsnm{Armon}, \binits{A.}}:
\batitle{Tabular data: Deep learning is not all you need}.
\bjtitle{Information Fusion}
\bvolume{81},
\bfpage{84}--\blpage{90}
(\byear{2022})
\doiurl{10.1016/j.inffus.2021.11.011}
\end{barticle}
\endbibitem

%%% 25
\bibitem[\protect\citeauthoryear{Pedregosa et~al.}{2011}]{scikit-learn}
\begin{barticle}
\bauthor{\bsnm{Pedregosa}, \binits{F.}},
\bauthor{\bsnm{Varoquaux}, \binits{G.}},
\bauthor{\bsnm{Gramfort}, \binits{A.}},
\bauthor{\bsnm{Michel}, \binits{V.}},
\bauthor{\bsnm{Thirion}, \binits{B.}},
\bauthor{\bsnm{Grisel}, \binits{O.}},
\bauthor{\bsnm{Blondel}, \binits{M.}},
\bauthor{\bsnm{Prettenhofer}, \binits{P.}},
\bauthor{\bsnm{Weiss}, \binits{R.}},
\bauthor{\bsnm{Dubourg}, \binits{V.}},
\bauthor{\bsnm{Vanderplas}, \binits{J.}},
\bauthor{\bsnm{Passos}, \binits{A.}},
\bauthor{\bsnm{Cournapeau}, \binits{D.}},
\bauthor{\bsnm{Brucher}, \binits{M.}},
\bauthor{\bsnm{Perrot}, \binits{M.}},
\bauthor{\bsnm{Duchesnay}, \binits{E.}}:
\batitle{Scikit-learn: Machine learning in {P}ython}.
\bjtitle{Journal of Machine Learning Research}
\bvolume{12},
\bfpage{2825}--\blpage{2830}
(\byear{2011})
{\href{https://arxiv.org/abs/1201.0490}{{1201.0490}}}
\end{barticle}
\endbibitem

%%% 26
\bibitem[\protect\citeauthoryear{Davis and Goadrich}{2006}]{auroc}
\begin{bchapter}
\bauthor{\bsnm{Davis}, \binits{J.}},
\bauthor{\bsnm{Goadrich}, \binits{M.}}:
\bctitle{The relationship between precision-recall and {ROC} curves}.
In: \bbtitle{Proceedings of the 23rd International Conference on Machine
  Learning}.
\bsertitle{ICML '06},
pp. \bfpage{233}--\blpage{240}.
\bpublisher{Association for Computing Machinery},
\blocation{New York, NY, USA}
(\byear{2006}).
\doiurl{10.1145/1143844.1143874}
\end{bchapter}
\endbibitem

\end{thebibliography}
